\documentclass[10pt,twocolumn,letterpaper]{article}

\usepackage{wacv}
\usepackage{times}
\usepackage{epsfig}
\usepackage{graphicx}
\usepackage{amsmath}
\usepackage{amssymb}
\usepackage{subcaption}
\usepackage{adjustbox}
\usepackage{multirow}

\usepackage{soul}
\newcommand{\myunderline}[2][black]{\setulcolor{#1}\ul{#2}\setulcolor{black}}

\def\wacvPaperID{1006} 

\wacvfinalcopy 

\ifwacvfinal
\def\assignedStartPage{9876} 
\fi

\ifwacvfinal
\usepackage[breaklinks=true,bookmarks=false]{hyperref}
\else
\usepackage[pagebackref=true,breaklinks=true,colorlinks,bookmarks=false]{hyperref}
\fi

\ifwacvfinal
\else
\fi

\begin{document}

\title{
Pushing the Envelope of Thin Crack Detection}

\author{Liang Xu$^{1}$
~~~~Taro Hatsutani$^{1}$
~~~~Xing Liu$^{1}$
~~~~Engkarat Techapanurak$^{1}$
~~~~Han Zou$^{1,2}$
~~~~Takayuki Okatani$^{1,2}$
\\
$^{1}$Graduate School of Information Sciences, Tohoku University
~~~~$^{2}$RIKEN Center for AIP \\
{\tt\small \{xu, hatsutani, ryu, engkarat, hzou, okatani\}@vision.is.tohoku.ac.jp}
}

\maketitle

\begin{abstract}
In this study, we consider the problem of detecting cracks from the image of a concrete surface for automated inspection of infrastructure, such as bridges. Its overall accuracy is determined by how accurately thin cracks with sub-pixel widths can be detected. Our interest is in making it possible to detect cracks close to the limit of thinness if it can be defined. Toward this end, we first propose a method for training a CNN to make it detect cracks more accurately than humans while training them on human-annotated labels. To achieve this seemingly impossible goal, we intentionally lower the spatial resolution of input images while maintaining that of their labels when training a CNN. This makes it possible to annotate cracks that are too thin for humans to detect, which we call {\em super-human labels}. We experimentally show that this makes it possible to detect cracks from an image of one-third the resolution of images used for annotation with about the same accuracy. We additionally propose three methods for further improving the detection accuracy of thin cracks:
i) $\mathcal{P}$-pooling to maintain small image structures during downsampling operations; ii) 
Removal of short-segment cracks in a post-processing step utilizing a prior of crack shapes learned using the VAE-GAN framework; iii) 
Modeling uncertainty of the prediction to better handle hard labels beyond the limit of CNNs' detection ability, which technically work as noisy labels.
We experimentally examine the effectiveness of these methods. 

\end{abstract}


\section{Introduction}


Automating infrastructure inspection such as roads, bridges, tunnels, etc. has recently attracted a lot of attention. One of the important tasks comprising the inspection is detecting cracks emerging on concrete surfaces of bridges, etc. Their images can be efficiently captured by a robot, e.g., an UAV, from which the existence of cracks can be automatically identified by computer vision methods. 


In this study, we consider the problem of detecting cracks from the image of a concrete surface. There are already a large number of studies of crack detection \cite{CrackForest,STRUM_AdaBoost,CrackTree206}. Recent studies employ convolutional neural networks (CNNs), having taken detection accuracy to a higher level. 
In the problem of crack detection, thick cracks are easy to detect while thin cracks are not. Thus, accuracy is mostly determined by how well the method can detect thin cracks.

It is practically important to accurately detect thin cracks for better management of infrastructure. In the case of concrete surfaces of a typical type of bridges, it is requested to find cracks with 0.5mm width and sometimes even those with 0.1mm width.  From a computer vision point of view, what is essential is the width of imaged cracks than the real widths, as they depend on the distance between the camera and the concrete surface. 

Humans and CNNs can recognize thin cracks with subpixel widths, owing to the large image structure along cracks. That said, there has to be a limit in the width of recognizable cracks. In this study, we are interested in making it possible to detect cracks close to the limit of thinness. 
Toward this ends, we propose a collection of techniques. 

\begin{figure}[t]
\centering
\includegraphics[width=0.65\columnwidth]{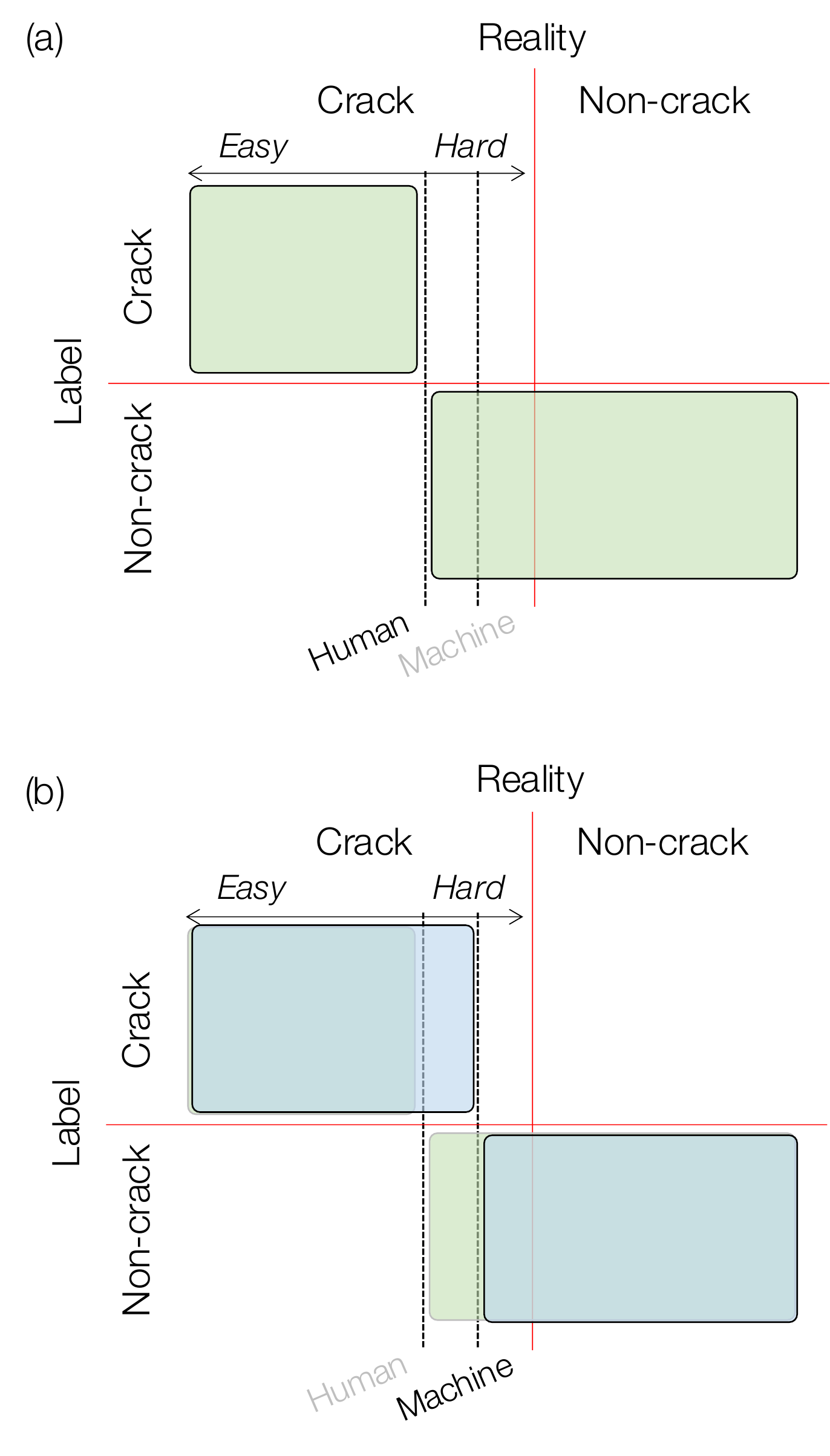}
\vspace*{-3mm}
\caption{(a) Human annotators can only label cracks that they can identify in images. The rest of the real cracks are labeled as non-crack. (b) A superior machine detector can detect cracks that are harder to detect. To train it properly, its labels should be created accordingly; the label boundary should be moved toward harder cracks, and ideally, it should coincide with the detector's performance limit. }
\label{fig:ch1_super_human_label}
\end{figure}

We first present a method that trains a CNN with {\em super-human} labels. Its aim is to make CNNs detect cracks more accurately than humans while training them on the labels annotated by humans. As shown in Fig.~\ref{fig:ch1_super_human_label}(a), humans cannot detect some of thinnest cracks from its image. When we train a CNN using human-annotated labels, the best outcome will be human annotation reproduction. Obviously, this will not make a machine detector surpassing the annotator(s) in detection accuracy. To overcome this, we intentionally lower the spatial resolution of input images while maintaining that of their labels when training a CNN. This enables the annotation of cracks that are too thin for humans to detect, which we call super-human labels. The details are given in Sec.~\ref{sec:super-human}.

We additionally propose several methods for further improving the accuracy of detecting thin cracks. To be specific, we propose to employ i) $\mathcal{P}$-pooling \cite{p_pooling} in the design of CNNs to extract small image structures; ii) removal of fake, short-segment cracks in a post-processing step utilizing a prior of crack shapes learned using the VAE-GAN framework \cite{vae-gan,scn}; iii) modeling uncertainty of the prediction to better handle hard labels beyond the limit of CNNs' detection ability, which technically work as noisy labels.

\section{Related Work}
\label{sec:ch2_related_work}
Detecting cracks emerging on the surface of various materials (e.g., roads, concrete surfaces, metals, etc.) from its image has attracted a lot of attention, because of its wide range of industrial applications. 
In early studies, image processing methods have been employed, such as thresholding~\cite{local_threshold_seg}, edge detectors~\cite{Canny, improved_canny}, ridge detectors~\cite{ridge_detector}, etc. CrackIT~\cite{CrackIT} 
is a comprehensive crack-detection approach by using traditional low-level image processing techniques. MFCD~\cite{MFCD} formulates  pavement crack detection as a multiscale image fusion problem based on the assumption that the cracks have lower intensity values than the background.
These approaches tend to be sensitive to noise, variance of brightness and contrast etc. or rely too much on assumptions on the appearance of cracks, limiting the applicability. 


More recently, machine-learning-based approaches were studied. In CrackForest~\cite{CrackForest}, a random structured forest are employed to find the mapping between crack patches and structure tokens, in which the paired ground-truths are necessary for the training. In \cite{AdaBoost}, textural descriptors are chosen by AdaBoost to better describe imaged cracks, and morphological filters are used to reduce the pixel intensity variance. The STRUM (a spatially tuned robust multifeature) AdaBoost \cite{STRUM_AdaBoost} is developed to classify the status of cracks in image patches. 
Although these methods perform better than earlier efforts based on pure image processing methods, detection accuracy is not enough for practical applications, arguably due to limited performance of feature extraction. 

As with many other visual recognition tasks, deep learning has contributed to significant performance improvement. Several studies  \cite{FFA,LPP,block_centre_pixel,Block_wise,cha_2017,Yang_2018AutomaticPC,deepcrack_detection_CrackTree260,Fist_road_crack_del} extract features from a patch and determine whether its center pixel is crack or not. They achieved better performance with transitional areas (i.e., the pixels on the crack/non-crack boundary) at the expense of computational cost. 
In \cite{zhang2018a,zhang2018b}, transfer learning is employed to use a CNN to classify pavement images into cracks and sealed-cracks; the method can detect hairline cracks while eliminating local noises and maintaining fast processing speed. In \cite{two_step}, a two-step approach is proposed to detect cracks on road pavement,  which first segment the target area from input image and then detect cracks from target area.
In \cite{huang_2018,bang_2019}, the standard formulation of semantic segmentation is employed, and popular CNN architectures (fully convolutional networks etc.) are adopted, making it possible to deal with images of different sizes. Currently, this formulation is the most popular 
for crack detection. The success of this approach mostly relies on the availability of good datasets that are given accurate annotation of cracks. It is not so easy to obtain/create such good datasets, limiting its applicability to the real world. 

Any visual recognition/detection tasks generally require a large-scale dataset to train a CNN, so does crack detection. There are several datasets for crack detection that are publicly available. For instance, expanded from an early CrackTree206 dataset~\cite{CrackTree206}, CrackTree260~\cite{cracktree} contains 260 images of road pavement, each of which is of 800$\times$600 pixel size. DeepCrack~\cite{deepcrack_segementation} is a public benchmark dataset containing 537 images of cracks in multiple scales and scenes that can be used for evaluation of crack detection methods. 
These datasets basically consists of images containing only thick cracks, and are not fit for the purpose of this study.

\section{The Problem: Crack Detection}

\subsection{Practical Requirements for Crack Detection}

Recent studies of crack detection tend to formulate crack detection as image segmentation. Their output is a pixel-wise crack map; each pixel has the likelihood indicating that the pixel lies on a crack. In this study, employing this formulation, we consider only two classes, i.e., crack and non-crack (i.e., background); then, it turns to a binary class segmentation problem. 

However, from a practical perspective, identifying cracks in a pixel-wise fashion is usually unnecessary. What is more important is determining the existence/nonexistence of crack(s) as precisely as possible. The detected crack's position does not need to be very accurate; a few pixel shifts from the ground truth position are perfectly allowable. Thus, it is sufficient to represent each crack as a one-dimensional curve, or a poly-line, with one-pixel width. Based on this requirement, we reconsider how to annotate cracks and evaluate their detectors. 

\subsection{Annotation of Cracks}
\label{sec:annotation}

Many studies follow the standard method for semantic segmentation for annotation of cracks, where cracks are annotated as pixel regions. This may be appropriate when the imaged cracks have widths of at least a few pixels. However, we are interested in detecting cracks, including those with sub-pixel widths; they do not form a `region.' Besides, the pixel-wise annotation is not efficient; it will take a long time annotate a single image. 

Considering also the above practical requirement, we employ a different approach for annotation. We represent a crack as a one-dimensional structure, or more precisely, as a set of connected pixels with the chain code. We developed an assisting tool for the annotation, using which the annotator needs only to click the start and end points of each crack. Then, the tool finds the path along the crack connecting the specified two points. It is based on an algorithm finding the shortest distance path, in which the distance traversing each pixel is designed to reflect its `crack-ness', which is calculated by a ridge detector \cite{ridge_detector}. 

\subsection{Training with One-pixel-width Crack Label}
\label{sec:one-pixel}

The choice of this annotation method poses two problems. One is how to fill in the differences between the images and the label maps when training our network. As we employ the same formulation as semantic segmentation, the network outputs a map of crack likelihood. Even for a thick crack having several pixel widths on the image, it has one-pixel width in the label map, resulting in that many pixels on the imaged crack will be annotated as {\em non-crack}. The standard loss function used for semantic segmentation may not work well in the presence of such label differences. 



We apply a Gaussian filter to the label maps to cope with this. A similar approach has been employed for human pose estimations (e.g., \cite{Hourglass}). In relation to this, we employ a mean squared error (MSE) loss for the loss function in the training of our CNNs instead of the standard binary cross-entropy (BCE) loss. We have empirically found that the former works better than the latter. The details are given in the supplementary.



\subsection{Evaluation of Crack Detection}
\label{sec:ch3_def_TP}

The other issue with the above annotation is how to evaluate detection results. Our network outputs a two-dimensional map of crack likelihood. By thresholding it, we get a binary crack prediction map. Assuming the ground truth map of crack labels, all the pixels are classified into true positive (TP), true negative (TN), false positive (FP), and false negative (FN). Then, the standard evaluation metrics for semantic segmentation, such as IoU and DICE coefficients, can be calculated by counting the numbers of these classified pixels. However, these metrics are not fit for our purpose, as the ground truth map has only one-dimensional crack labels whereas the predicted cracks tend to have a few pixel widths, partly due to the fundamental difficulty with predicting segmented `regions' with one pixel width and partly due to the label smoothing explained above.

We dilate the crack labels with a small disk when calculating TPs and FPs to address this issue. If the predicted crack pixels are inside the dilated region, they are counted as TPs, and if not, the pixels are counted as FPs. This method avoids penalizing predicted crack pixels that do not precisely lie on the true crack. It also absorbs slight positional errors of predicted cracks; they are totally allowable in practice, as mentioned earlier. The radius of the dilating disk should desirably be the smallest that achieves these effects, to make it possible to distinguish two isolated cracks that are close to each other. We set the radius to three pixels in our experiments. We use the original one-pixel-width label when counting TNs and FNs. 

\section{Methods}

\subsection{Pseudo Super-human Labels by Lowering Image Resolution}
\label{sec:super-human}

As shown in Fig.~\ref{fig:ch1_super_human_label}(a), humans cannot detect all the cracks emerging on a concrete surface from its image. When we train a CNN using human-annotated labels, the best outcome is the human annotation's perfect reproduction. Is it possible to break this limit?

Let us assume that a CNN can potentially detect thinner and thus harder cracks than humans, as shown in Fig.~\ref{fig:ch1_super_human_label}(b). To train the CNN to unleash its potential, different training data in which such hard cracks are given the positive label are necessary. Ideally, the annotation should desirably coincide with the upper bound of its potential performance. 

However, this is easier said than done. To annotate hard cracks in images as {\em crack} beyond humans' ability, we will need different sensor data to better identify real cracks. For example, we can capture additional images of the same concrete surface using a different camera with higher resolution or simply placing the original camera more closely to the surface. Even if having the same annotator annotate the new images, thinner cracks that are harder to detect will be correctly identified in these images, from which we can get proper labels for the original images.

\begin{figure}
    \centering
    \includegraphics[width=\linewidth]{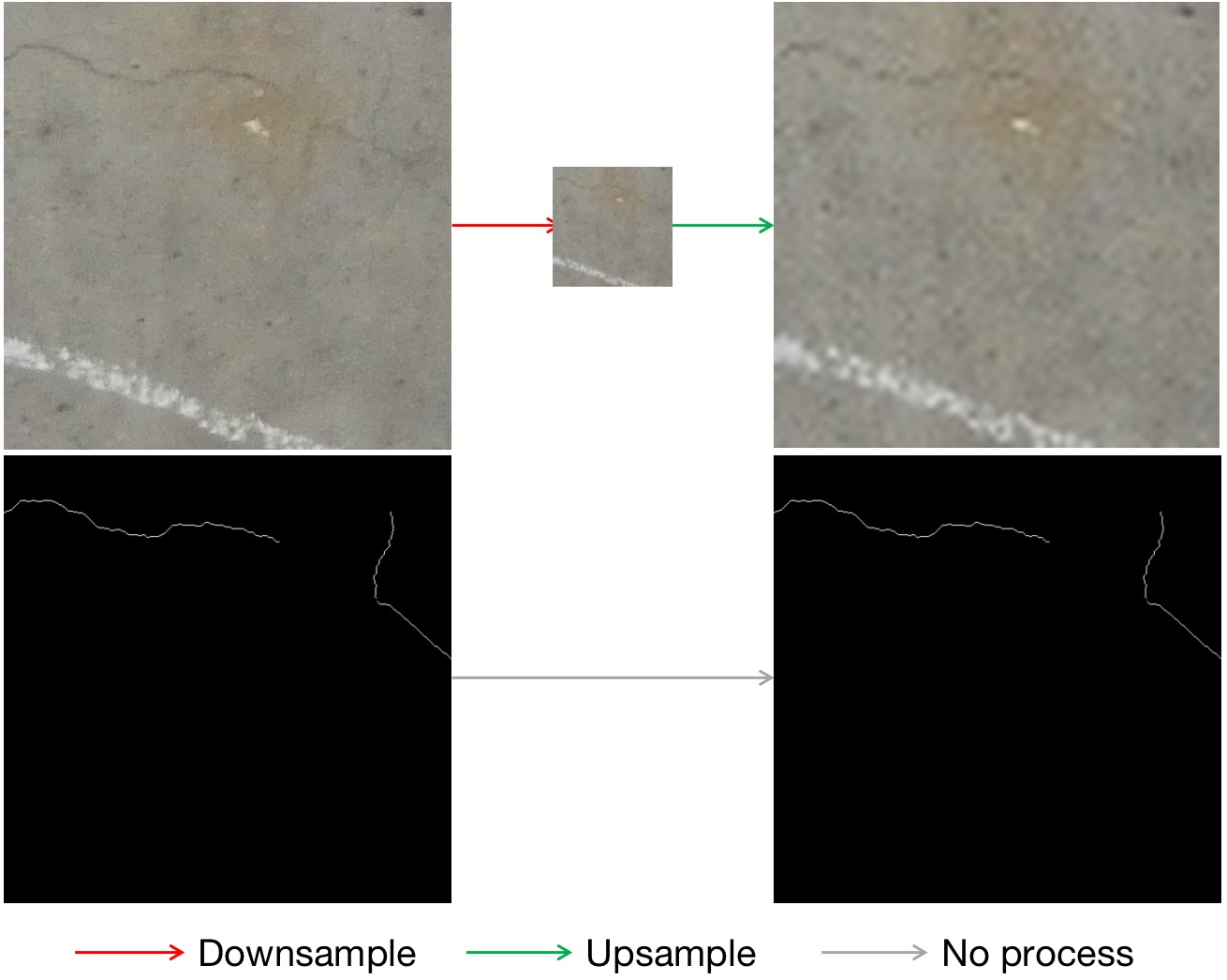}
    \caption{Lowering image resolution by downsampling an image and then upsampling the downsampled image with the same factor. The label map are kept unprocessed. The factor is four with the above example; we denote it by $\times 4$ (i.e., four times lower resolution). }
    \label{fig:ch4_down_up}
\end{figure}

The primary issue with this approach is its cost. Every image in the training data needs a corresponding higher-resolution image(s). We consider creating such pairs of low- and high-resolution images by image synthesis based on simple downsampling to cope with this. To be specific, assuming an image and a label annotated for it, we downsample the image while maintaining the label map `as is.' We then upsample the downsampled image to the original size, due to the reason explained below, and pair it with the original label map to create an image-label pair for training (and testing) (See Fig.~\ref{fig:ch4_down_up}.). 
Intuitively, high-resolution image is used as evidence for achieving pseudo super-human label, and a pair of a corresponding low-resolution image and a pseudo super-human label is for training (and testing).

This down-/up-sampling processing may lead to undetectable cracks which means that some thin cracks in row images disappear in newly created images. A example is showed in Fig.~\ref{fig:ch4_down_up}. A strength of this approach is that we can control the boundary between detectable and undetectable cracks shown in Fig.~\ref{fig:ch1_super_human_label}(a) and (b) by changing the downsampling factor. We conduct experiments in which we vary the downsampling factor to create training/test data to study the relation of the limit of human annotation and that of the CNN detectors. 

We employ the pair of image down- and up-sampling, as mentioned above, to precisely evaluate the effects of different downsampling factors. For instance, $\times 1/8$ downsampling of an image shrinks the image area to $1/64$. This will lead to severe imbalance of data size, as  the amount of training data is proportional to the number of image pixels in the case of pixel-wise classification (i.e., segmentation tasks). 

\subsection{Methods for Further Improvements} 
\label{sec:ch4_improve_methods}

We consider several additional methods to improve the accuracy of detecting hard cracks. 

\subsubsection{Pooling Sensitive to Spatial Input Perturbation}


In \cite{p_pooling}, a new pooling method, named $\mathcal{P}$-pooling, is proposed, aiming at equipping the pooling operation with both sensitivity to spatial input perturbation and non-linearity of the computation. It is defined by
\begin{equation}
    f(R_{j})=\frac{\sum _{x_{j}\subseteq R_{j}}x_{j}^{p+1}}{\sum _{x_{j}\subseteq R_{j}}x_{j}^{p}},
    \label{equation:ch4_p-pooling}
\end{equation}
where $R_j$ is the $j$-th pooling region in the input feature map and $p$ is a hyperparameter or a learnable parameter. The standard max pooling has the latter but not the former, while the average pooling has the former but not the latter. The study's original motivation is to better use a pre-trained CNN on a classification task for segmentation tasks. The two tasks have opposite objectives regarding the (in)sensitivity to spatial input perturbation. The study aims at reconciling them by replacing the pooling/downsampling operation in the pre-trained CNN with $\mathcal{P}$-pooling. 

Although our CNNs do not rely on pre-training, we have found that the employment of $\mathcal{P}$-pooling improves the accuracy of crack detection. This is arguably due to the increased sensitivity to input perturbation; it is crucial for detecting thin cracks. 
We replace all the max-pooling layers in the U-Net architecture with this pooling method. 




\subsubsection{Eliminating Short Segments by Shape Prior}

Cracks emerging on concrete surfaces tend to have more than a certain length and usually do not form isolated short segments due to a physical reason. As a matter of fact, this is also the case with annotated label maps. Our CNNs are trained on them, and thus ideally, they should not produce isolated short-segment cracks. Although this is true when there are only easy (i.e., thick) cracks in the training/testing data, it is not otherwise. When the training data contains hard cracks, and we train our CNNs to detect them, it starts to produce short segment cracks in the predicted map. This happens, probably because the labels beyond the performance limit of CNNs may technically work as noisy labels. 

In any case, we need additional treatment to deal with the emergence of such short-segment cracks. We propose to learn a prior on the shape of cracks and use it to eliminate them in a post-processing stage. Measuring the length of predicted cracks with image processing methods such as finding connected components will not work, since it is difficult to distinguish isolated short segments from a series of short segments forming a long crack. 

To cope with this, we use an auto-encoder to learn the shape prior of cracks. Several attempts \cite{medica_image_seg,shape-aware_organ_seg} were proposed in the past targeted the segmentation of biological organs from medical images. We adopt a method (shape constrained network; SCN) proposed in \cite{scn} that utilizes the VAE-GAN~\cite{vae-gan}. Crack shapes are arbitrary, and their variety will be larger than the shape of an eye considered in \cite{scn}. Employing the same network architecture as \cite{scn}, we test the effectiveness of this method in our experiments. 



\begin{figure}
    \centering
    \begin{subfigure}{1.0\linewidth}
        \centering
        \includegraphics[trim={60 0 30 0}, clip, width=\linewidth]{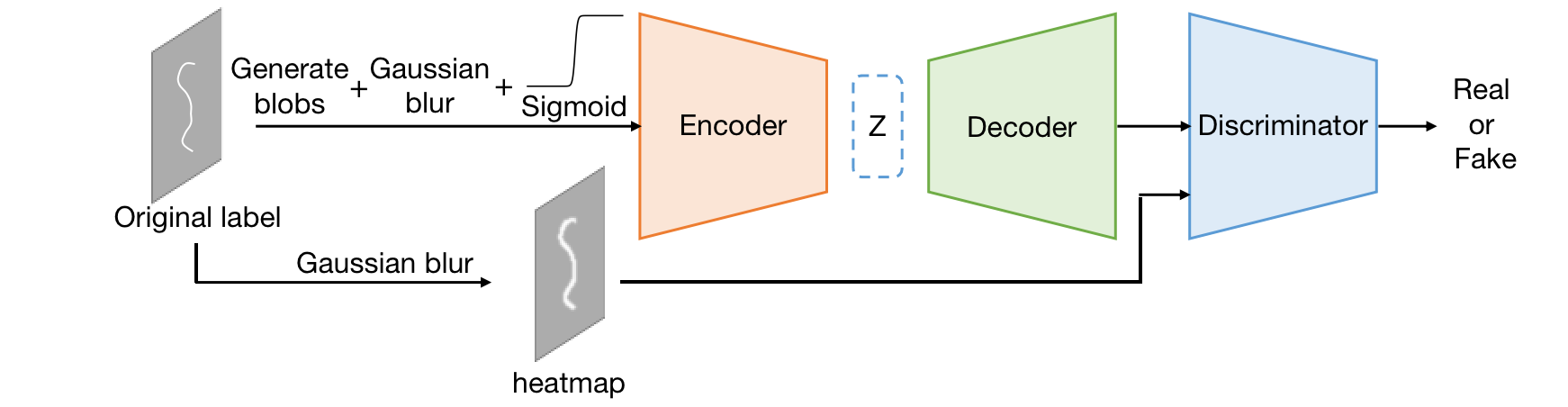}
        \caption{}
    \end{subfigure}\\
    \begin{subfigure}{1.0\linewidth}
        \centering
         \includegraphics[trim={60 0 30 0}, clip, width=\linewidth]{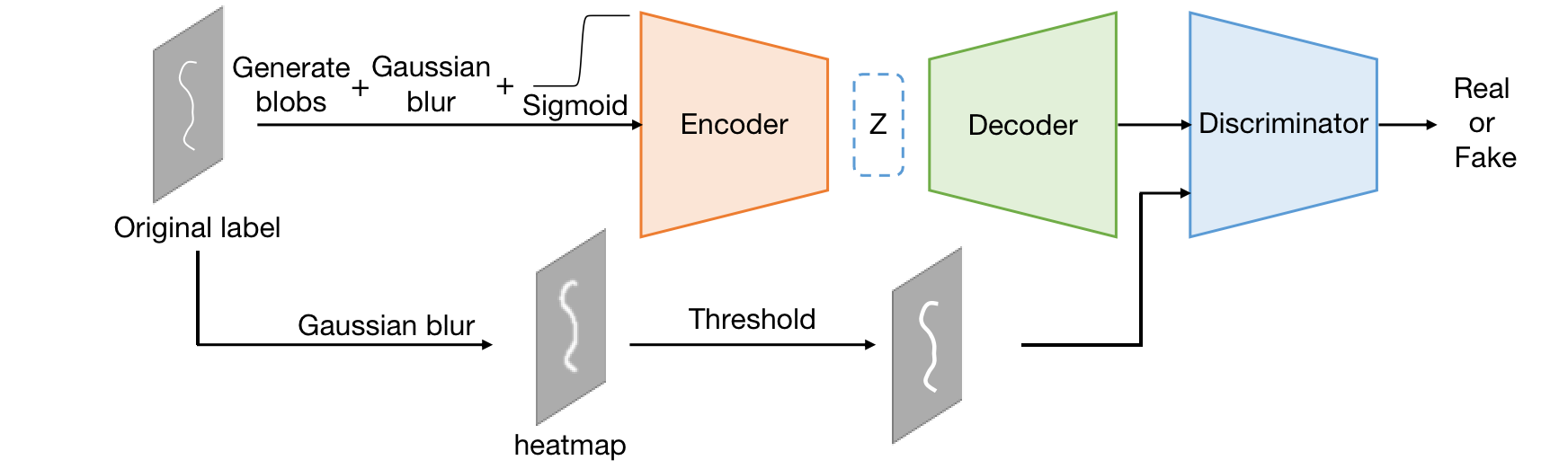}
         \caption{}
    \end{subfigure}
\caption{Tested two training methods of shape constrained network (SCN) to remove short segments of cracks in the predicted map.}
\label{fig:ch4_vae_train}
\end{figure}

In our experiments, we test two methods shown in Fig.~\ref{fig:ch4_vae_train}. To make the VAE learn to remove short segments, which tend to have the shape of blobs in the predicted maps, we randomly generate several oval blobs in the input label maps. We then apply Gaussian blur to the maps, followed by the application of a logistic sigmoid function. The resulting images are inputted to the VAE. 
The sigmoid function is used as a proxy of binarization that is differentiable:
\begin{equation}\nonumber
    y = \mathrm{logistic}(K x - I_{\mathrm{th}}),
\end{equation}
where we set $K=20$ and $I_{\mathrm{th}}=0.4$ in our experiments.
We train the VAE together with a discriminator classifying its input as the ground truth label map (blurred and optionally binarized) or the generated one in the framework of VAE-GAN.


\subsubsection{Consideration of Prediction Uncertainty}

Cracks lying around the limit of humans' detection ability can be annotated erroneously. Furthermore, for cracks beyond the performance limit of CNNs, their correct labels may technically work as noise labels, as shown in Fig.~\ref{fig:ch1_super_human_label}. To cope with such wrongly annotated or too hard labels, we consider modeling uncertainty of the prediction \cite{uncertainty}. As mentioned above, we use an MSE loss $(y-\overline{y})^2$ for training the CNNs due to its good performance, where $y$ is the predicted crack likelihood and $\overline{y}\in[0,1]$ is the Gaussian blurred label. Regarding $\overline{y}$ to be an observation $N(\mu,\sigma^2)$. Then, training with only an MSE loss is equivalent to estimating $\mu$ while setting $\sigma$ to be a constant. Estimating not only $\mu$ but $\sigma^2$ in the framework of maximum likelihood, the loss function turns to
\begin{equation}
   {\frac{(\overline{y}-\mu )^{2}}{2\sigma ^{2}}}+\frac{1}{2}\log\sigma^{2}.
\end{equation}
We reparametrize as $s\equiv \log\sigma^{2}$. We add a single unit to the final output layer of the CNN to predict $s$. The only unit already existing in the original output layer predicts $\mu$. Thus, the loss for the two output $(\mu, s)$ is given by
\begin{equation}
    {\frac{(\mu - \bar{y} )^{2}}{2e^{s}}}+\frac{1}{2}s.
    \label{eqn:uncertainty_loss}
\end{equation}

The above loss makes it possible to predict the uncertainty of the crack prediction. When minimizing the loss, if the CNN can predict $\mu$ that is sufficiently close to the label $\overline{y}$, then $\sigma^2$ will be small. If it cannot, then $\sigma^2$ will be large to compensate for the squared error. Thus, the CNN learns to predict $\sigma$ depending on the certainty of the prediction of $\mu$ \cite{uncertainty}. However, we are not interested in this predictive uncertainty itself here. We rather wish that this mechanism will help handle noisy labels, e.g., when an invisible crack is annotated as {\em crack}.  

\section{Experimental Results}

\subsection{Experimental Configuration}
\label{sec:ch5_config}

\paragraph{Dataset}

\label{sec:ch3_labeling}

As mentioned in Sec.~\ref{sec:ch2_related_work}, there is no dataset fit for the purpose of this study. Thus, we create the dataset by ourselves. It contains 352 images of concrete surfaces of various bridges captured by hand-held cameras. Their sizes are in the range from 4,000$\times$ 3,000 to 5,000$\times$ 4,000 pixels. Examples are provided in the supplementary. 
We had three persons annotate these images by the semi-automatic method, which produces crack labels with one-pixel width, as explained in Sec.~\ref{sec:annotation}. We split the images into 282 for training/validation and 70 for test. 
Our original images are too large for CNNs. Thus, we crop into small patches (e.g.512$\times$512). To increase the size of training dataset,  we crop patches from images by using a sliding window with a half size of the patch size. Thus, we achieve 24700 patches (512$\times$512) for training, and 4972 patches for testing.


\paragraph{Network architecture}

We choose U-Net \cite{U_net}
for the baseline architecture of CNNs. 
Our preliminary experiments have found that a U-Net variant with narrow layer widths works fairly well. 
Thus, we first reduce the number of channels of every convolutional layer to 1/4, setting them to be \{1-16-32-64-128-256-128-64-32-16-1\}. This decreases the number of trainable parameters from around 31.0 million to only 1.9 million, and also contributes significantly faster training. We then add a batch normalization layer after every convolutional layer (before activation layer), which contributes to stabilization of training. This network shows the top-level accuracy in our experiments. We call this U-Net*+BN later. 

A small but noticeable gain is further obtained by modification of the network design. We experimentally tested several modifications, such as incorporation of the residual (Res) blocks \cite{ResNet} and its variants as well as the squeeze-and excitation (SE) block \cite{SE_block} and its variants. From the results, we choose the model in which every convolutional layer of the above base network is replaced with a modified Res block with an additional standard SE block. The details of the comparisons are given in the supplementary.


\paragraph{Training methods}

As the images in our dataset are large, we crop patches of a fixed size and feed them to our CNNs; $572\times 572$  for U-Net*+BN and $512\times 512$ for the others. We crop patches in a sliding window fashion with a stride of a half size of the patch size. 
There are much more negative patches (patches without any crack pixel) than positive patches (patches with crack pixels). We employ random minority oversampling for selection of patches used for training. The patches are augmented by random 360-degree rotation and random flipping.

Unless otherwise noted, all models are trained by the Adam optimizer~\cite{Adam} with initial learning rate of $1e^{-3}$ , $\beta _{1}$ = 0.9 and $\beta _{2}$ = 0.999. 
All the parameters in convolutional layers are initialized using the He initialization \cite{Delving}. We train the model for 60 epochs and drop the learning rate to $1e^{-4}$ after 50 epochs. We set the size of minibatches to 16. 
We use PyTorch~\cite{Pytorch} for the experiments. 
 
\begin{figure*}[t]
\hfil
$\begin{array}{ccc}
\includegraphics[width=0.32\textwidth]{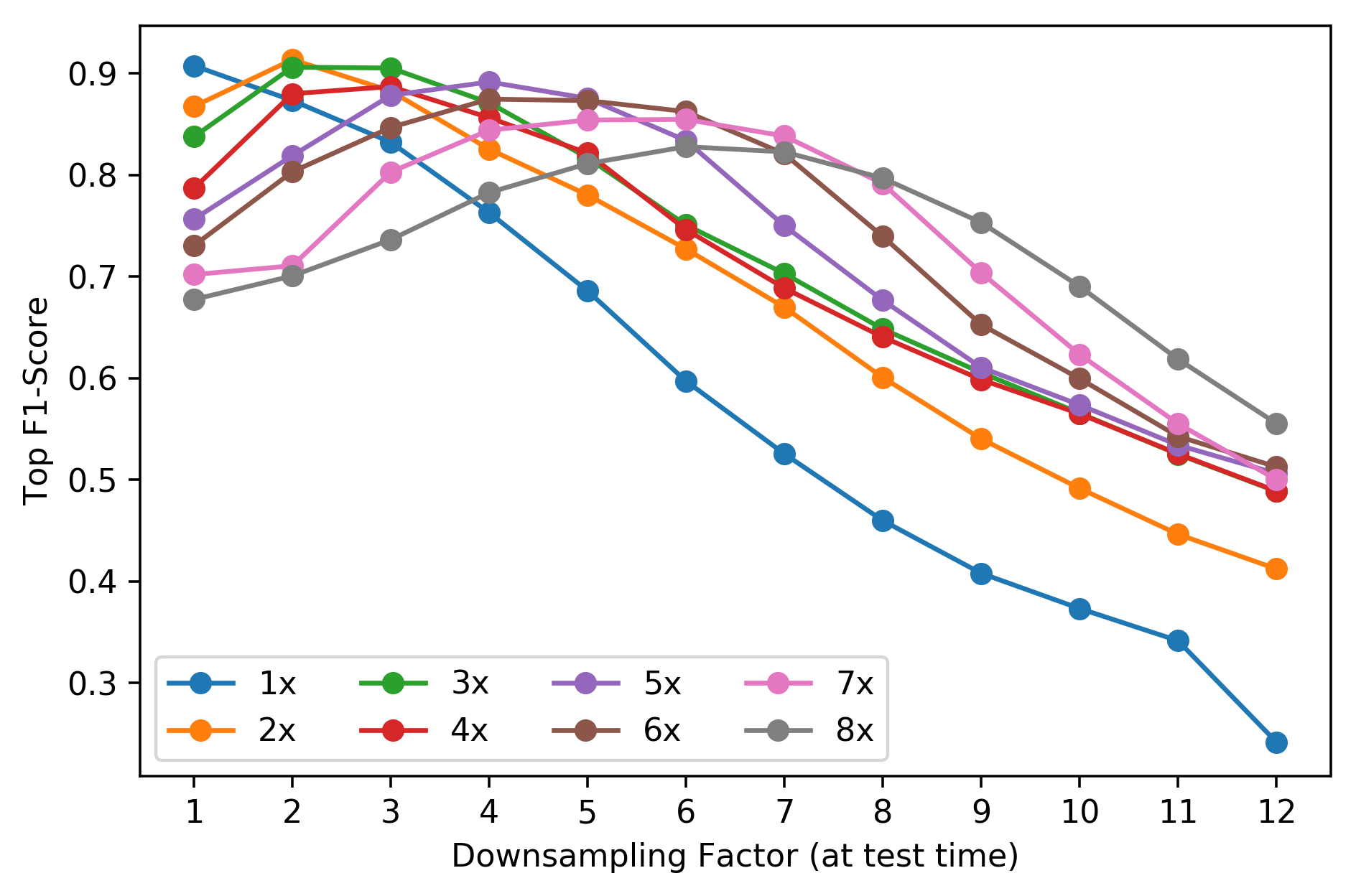}\!\!&
\includegraphics[width=0.32\textwidth]{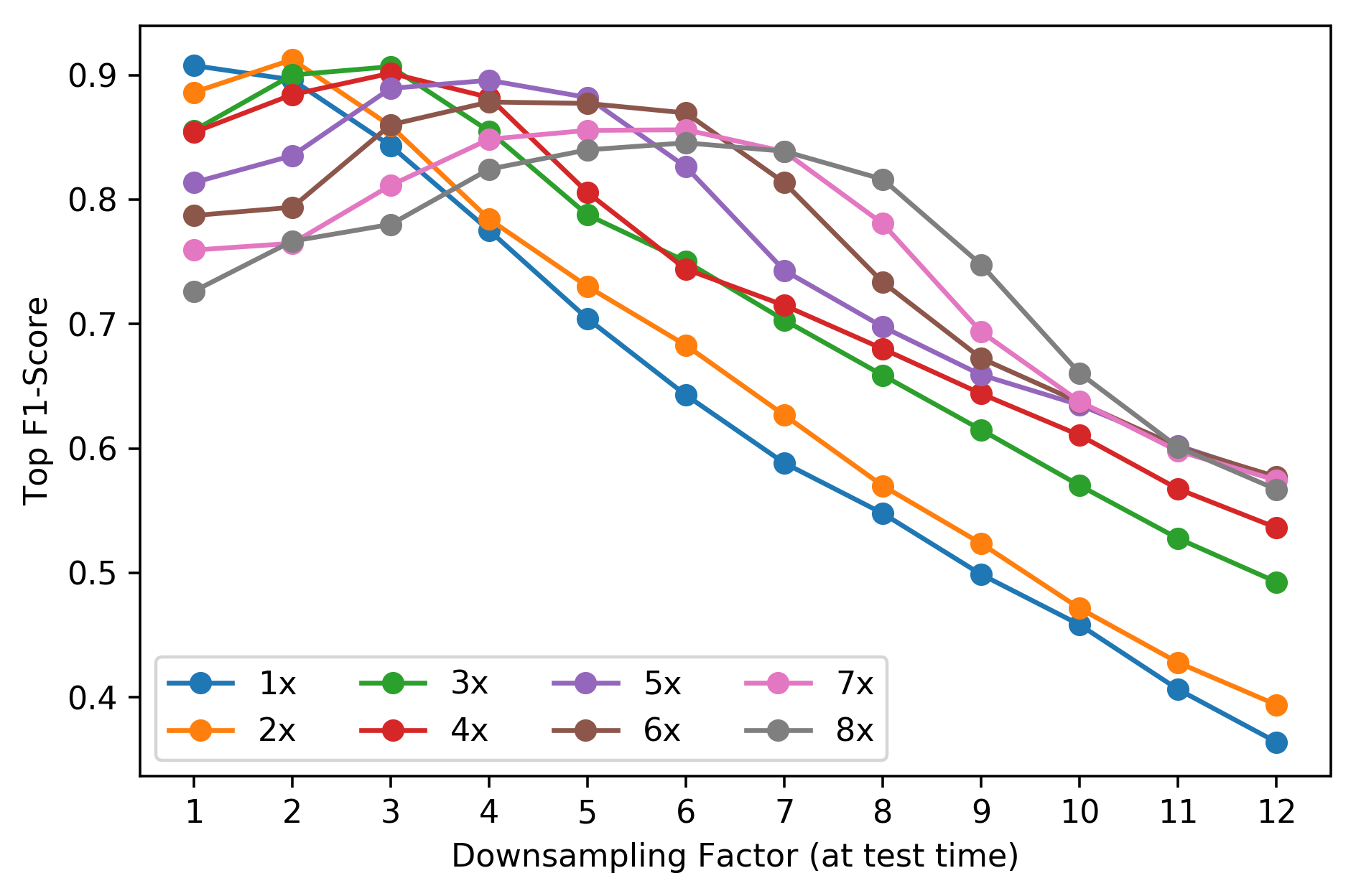}\!\!&
\includegraphics[width=0.32\textwidth]{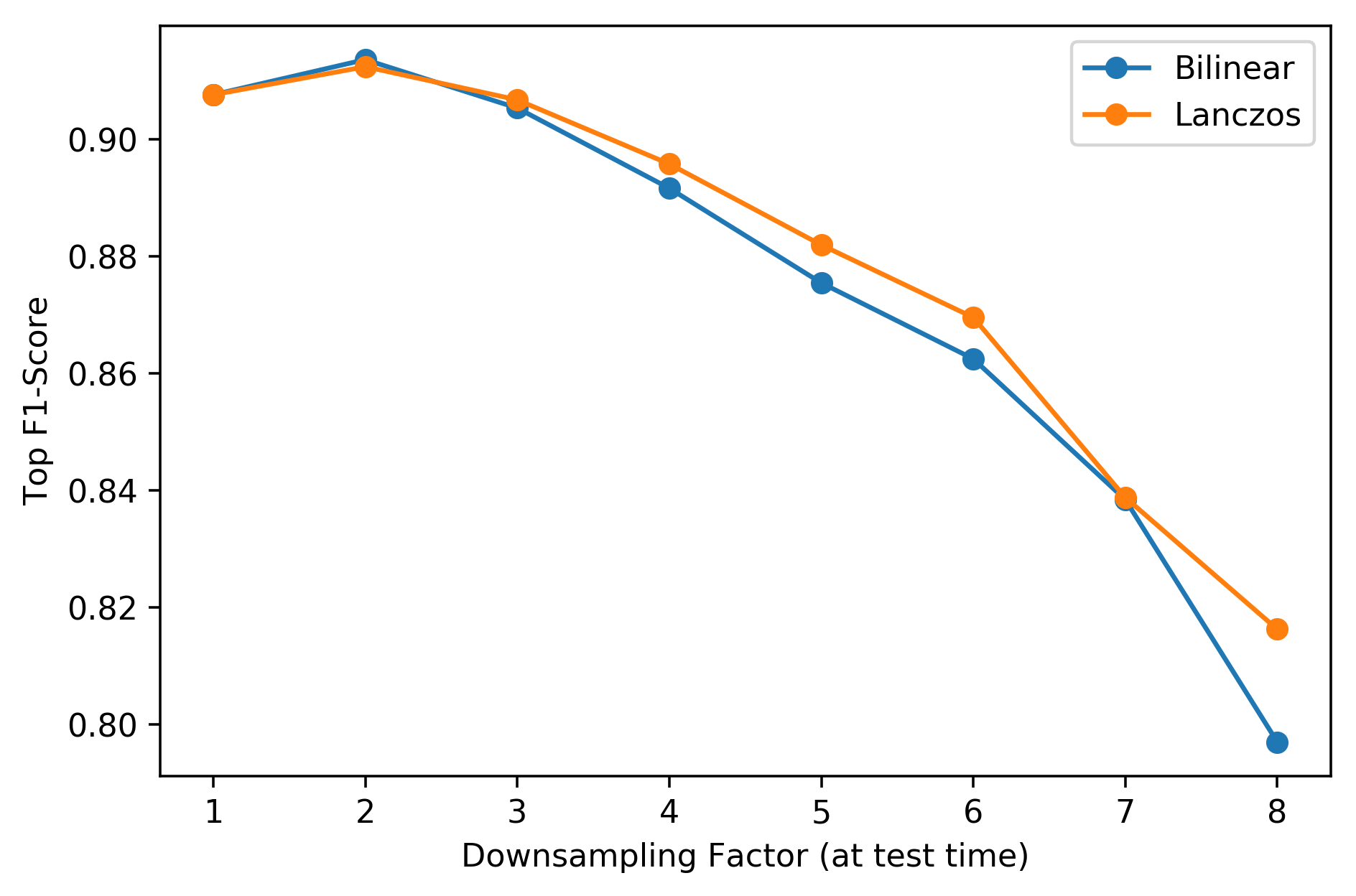}\\
\mbox{\footnotesize (a)}&\mbox{\footnotesize (b)}&\mbox{\footnotesize (c)}
\end{array}$
\caption{Detection accuracy (F1-score) of networks trained and tested on downsampled images. Each curve ($n\times$) indicates a network trained on images downsampled by $n:1$ followed by upsampling of $1:n$. The horizontal axis indicates the factor of downsampling (and the subsequent upsampling) applied to the test images. }
\label{fig:ch5_sampling_train_test}
\end{figure*}

\paragraph{Evaluation}
\label{sec:ch4_evaluation}





Our networks output a two-dimensional map storing crack likelihood at each pixel. We then bianarize the map with a threshold $=0.5$, and compute TP, FP, TN, and FN using the method explained in Sec.~\ref{sec:ch3_def_TP}. In all the experiments, we repeat each trial three times and report their average. 


\subsection{Effects of Training on Super-human Annotation}
\label{sec:ex_superhuman}

We first conducted an experiment to see the effects of training on down-/up-sampled images. For brevity, we refer to the factor of down-/up-sampling applied to images as {\em resolution}. Figure~\ref{fig:ch5_sampling_train_test}
shows the detection accuracy measured by F1-score for the models trained and tested on different resolutions. To be specific, we first train a network (U-Net*+BN) from scratch on the images down-/up-sampled by a factor of $n$, which is denoted by ``$n\times$'' in the figure. Each network is then tested on the images down-/up-sampled by a factor of $m$, which is specified by the horizontal axis of the plots. We do this training and test for every pair of $n\in[1,8]$ and $m\in[1,12]$.  Figures~\ref{fig:ch5_sampling_train_test}(a) and \ref{fig:ch5_sampling_train_test}(b) show the results obtained when bilinear interpolation and Lanczos interpolation are used for down-/up-sampling, respectively. Figure~\ref{fig:ch5_sampling_train_test}(c) shows the top `envelop' of the curves, which represents the best score for each resolution of test images. We can make several observations from these figures. 

It is first observed that the highest score at each test image resolution $m$ is achieved by the model trained on the same resolution ``$n\times$'' (except the model of ``$4\times$'', which achieves the best score at the test image resolution=``$3\times$''). This is reasonable since the potential domain shift between the training and test images should be the smallest when their resolutions are the same. 

It is also seen from Fig.~\ref{fig:ch5_sampling_train_test}(c) that the highest scores at test resolutions $m=1$, $2$, and $3$ are very close to each other. That of $m=2$ is even slightly higher than $m=1$. We think that these prove the effectiveness of the super-human labels. The test images of resolutions $m=2$ and $3$ contain cracks that are harder to detect due to their decreased resolution, to which nevertheless they are labeled as {\em crack}. This makes each model possible to learn to detect such hard cracks, achieving detection accuracy similar to resolution $m=1$. 

This result means that we can achieve (at least) the same accuracy level with up to $3\times$ lower resolution by using the proposed down-sampling method as the accuracy obtained by the standard training with the native resolution. Note that this is achieved using the labels annotated by humans on images with their native resolution. This has multiple practical implications. For instance, images captured from a $n\times$ distance to the concrete surface will have $n\times$ lower resolution. Thus, we can achieve the maximum accuracy for images captured from a distance three times farther from the target than the distance at which the annotated images are captured. 

It is seen from Fig.~\ref{fig:ch5_sampling_train_test}(c) that the detection accuracy decreases gradually but monotonically from resolution $m\geq 4$, which will be because it is beyond the detection ability of the CNN; cracks that are too hard for the CNN to detect are annotated as {\em crack}. The detection accuracy can decrease due to the following two reasons: i) hard cracks in the test images are simply too hard to detect, and ii) the annotated labels for the hard cracks work as noisy labels for the training, further lowering the accuracy. Further analyses are left to a future study. As for interpolation methods, the tendency is mostly similar but the results with the Lanczos interpolation are slightly better for the lower resolution cases.

\begin{table*}[thb]
    \footnotesize
    \centering
    \caption{Effectiveness of $\mathcal{P}$-pooling. All the pooling layers in U-Net*-BN are replaced with the pooling. Before/after the replacement. }
\resizebox{1.0\textwidth}{!}{
\begin{tabular}{c|c|cccccccc} 
\hline
\multicolumn{1}{c}{} & \multicolumn{1}{l|}{} & \multicolumn{8}{c}{Down-sampling factor (at test time)} \\ \cline{3-10}

\multicolumn{1}{l}{} & \multicolumn{1}{l|}{} & $\times$1 & $\times$2 & $\times$3 & $\times$4 & $\times$5 & $\times$6 & $\times$7 & $\times$8 \\ \cline{3-10}

\multicolumn{1}{c}{} & \multicolumn{1}{l|}{} & \multicolumn{8}{c}{\begin{tabular}[c]{@{}c@{}}F1-score: before / after\\
\end{tabular}} \\ \hline

\multirow{8}{*}{\begin{tabular}[c]{@{}c@{}}Down-sampling \\ factor\\ (at train time) \end{tabular}} & $\times$1 & 0.905 / \textbf{0.909} & 0.894 / \textbf{0.895} & 0.843 / \textbf{0.845} & 0.774 / \textbf{0.778} & 0.701 / \textbf{0.706} & 0.641 / \textbf{0.644} & 0.585 / \textbf{0.589} & 0.548 / \textbf{0.553}  \\ \cline{3-10}

& $\times$2 & 0.885 / \textbf{0.890}  & 0.913 / 0.913 & 0.857 / \textbf{0.860} & 0.783 / 0.781 & 0.718 / \textbf{0.723} & 0.681 / \textbf{0.686} & 0.624 / \textbf{0.631} & 0.570 / \textbf{0.577}  \\ \cline{3-10}

 & $\times$3 & 0.856 / \textbf{0.867}  & 0.901 / \textbf{0.907} & 0.908 / \textbf{0.910} & 0.855 / \textbf{0.859} & 0.787 / \textbf{0.793} & 0.750 / \textbf{0.757} & 0.702 / \textbf{0.710} & 0.659 / \textbf{0.669}  \\ \cline{3-10}

 & $\times$4 & 0.854 / \textbf{0.866}  & 0.882 / \textbf{0.889} & 0.902 / \textbf{0.904} & 0.882 / \textbf{0.885} & 0.804 / \textbf{0.810} & 0.742 / \textbf{0.750} & 0.701 / \textbf{0.721} & 0.679 / \textbf{0.692}  \\ \cline{3-10}

 & $\times$5 & 0.813 / \textbf{0.827}  & 0.834 / \textbf{0.846} & 0.889 / \textbf{0.893} & 0.895 / \textbf{0.897} & 0.881 / \textbf{0.886} & 0.827 / \textbf{0.833} & 0.743 / \textbf{0.757} & 0.695 / \textbf{0.710}  \\ \cline{3-10}

 & $\times$6 & 0.786 / \textbf{0.804}  & 0.794 / \textbf{0.811} & 0.853 / \textbf{0.864} & 0.877 / \textbf{0.883} & 0.876 / \textbf{0.880} & 0.866 / \textbf{0.872} & 0.813 / \textbf{0.819} & 0.734 / \textbf{0.752}  \\ \cline{3-10}

 & $\times$7 & 0.759 / \textbf{0.772}  & 0.763 / \textbf{0.789} & 0.811 / \textbf{0.835} & 0.847 / \textbf{0.861} & 0.855 / \textbf{0.863} & 0.856 / \textbf{0.863} & 0.839 / \textbf{0.847} & 0.778 / \textbf{0.791}  \\ \cline{3-10}

 & $\times$8 & 0.725 / \textbf{0.740}  & 0.765 / \textbf{0.786} & 0.779 / \textbf{0.803} & 0.825 / \textbf{0.836} & 0.840 / \textbf{0.847} & 0.846 / \textbf{0.853} & 0.839 / \textbf{0.845} & 0.816 / \textbf{0.829}  \\ \hline
\end{tabular}
}
    \label{tab:ch4_p-pooling_f1}
\end{table*}

\begin{table*}[thb]
    \footnotesize
    \centering
    \caption{Effectiveness of elimination of short segment cracks in a post-processing step. Before/after the application of the method. }

\resizebox{1.0\textwidth}{!}{
\begin{tabular}{c|c|cccccccc} 
\hline
\multicolumn{1}{c}{} & \multicolumn{1}{l|}{} & \multicolumn{8}{c}{Down-sampling factor (at test time)} \\ \cline{3-10}

\multicolumn{1}{l}{} & \multicolumn{1}{l|}{} & $\times$1 & $\times$2 & $\times$3 & $\times$4 & $\times$5 & $\times$6 & $\times$7 & $\times$8 \\ \cline{3-10}

\multicolumn{1}{c}{} & \multicolumn{1}{l|}{} & \multicolumn{8}{c}{\begin{tabular}[c]{@{}c@{}}F1-score: before / after
\end{tabular}} \\ \hline

\multirow{8}{*}{\begin{tabular}[c]{@{}c@{}}Down-sampling \\ factor\\ (at train time) \end{tabular}} & $\times$1 & 0.909 / \textbf{0.911} & 0.895 / \textbf{0.898} & 0.845 / \textbf{0.849} & 0.778 / \textbf{0.783} & 0.706 / \textbf{0.709} & 0.644 / \textbf{0.645} & 0.589 / \textbf{0.593} & 0.553 / \textbf{0.555}  \\ \cline{3-10}

& $\times$2 & 0.890 / \textbf{0.892}  & 0.913 / \textbf{0.917} & 0.860 / \textbf{0.864} & 0.781 / \textbf{0.784} & 0.723 / \textbf{0.725} & 0.686 / \textbf{0.688} & 0.631 / \textbf{0.634} & 0.577 / \textbf{0.580}  \\ \cline{3-10}

 & $\times$3 & 0.867 / \textbf{0.869}  & 0.907 / \textbf{0.910} & 0.910 / \textbf{0.916} & 0.859 / \textbf{0.864} & 0.793 / \textbf{0.796} & 0.757 / \textbf{0.760} & 0.710 / \textbf{0.714} & 0.669 / \textbf{0.671}  \\ \cline{3-10}

 & $\times$4 & 0.866 / \textbf{0.869}  & 0.889 / \textbf{0.894} & 0.904 / \textbf{0.909} & 0.885 / \textbf{0.895} & 0.810 / \textbf{0.814} & 0.750 / \textbf{0.752} & 0.721 / \textbf{0.726} & 0.692 / \textbf{0.694}  \\ \cline{3-10}

 & $\times$5 & 0.827 / \textbf{0.832}  & 0.846 / \textbf{0.849} & 0.893 / \textbf{0.898} & 0.897 / \textbf{0.904} & 0.886 / \textbf{0.899} & 0.833 / \textbf{0.843} & 0.757 / \textbf{0.762} & 0.710 / \textbf{0.714}  \\ \cline{3-10}

 & $\times$6 & 0.804 / \textbf{0.806}  & 0.811 / 0.811 & 0.864 / \textbf{0.871} & 0.883 / \textbf{0.891} & 0.880 / \textbf{0.893} & 0.872 / \textbf{0.887} & 0.819 / \textbf{0.835} & 0.752 / \textbf{0.762}  \\ \cline{3-10}

 & $\times$7 & 0.772 / \textbf{0.777}  & 0.789 / \textbf{0.792} & 0.835 / \textbf{0.841} & 0.861 / \textbf{0.866} & 0.863 / \textbf{0.870} & 0.863 / \textbf{0.878} & 0.847 / \textbf{0.866} & 0.791 / \textbf{0.807}  \\ \cline{3-10}

 & $\times$8 & 0.740 / \textbf{0.744}  & 0.786 / \textbf{0.788} & 0.803 / \textbf{0.809} & 0.836 / \textbf{0.841} & 0.847 / \textbf{0.858} & 0.853 / \textbf{0.867} & 0.845 / \textbf{0.865} & 0.829 / \textbf{0.856}  \\ \hline
\end{tabular}
}
    \label{tab:ch4_sr_f1}
\end{table*}

\begin{table*}[t]
    \footnotesize
    \centering
    \caption{Effectiveness of training with the uncertainty modeling. The standard MSE loss/the augmented loss with uncertainty. }

\resizebox{1.0\textwidth}{!}{
\begin{tabular}{c|c|cccccccc}
\hline
\multicolumn{1}{c}{} & \multicolumn{1}{l|}{} & \multicolumn{8}{c}{Down-sampling factor (at test time)} \\ \cline{3-10}

\multicolumn{1}{l}{} & \multicolumn{1}{l|}{} & $\times$1 & $\times$2 & $\times$3 & $\times$4 & $\times$5 & $\times$6 & $\times$7 & $\times$8 \\ \cline{3-10}

\multicolumn{1}{c}{} & \multicolumn{1}{l|}{} & \multicolumn{8}{c}{\begin{tabular}[c]{@{}c@{}}F1-score: before / after
\end{tabular}} \\ \hline

\multirow{8}{*}{\begin{tabular}[c]{@{}c@{}}Down-sampling \\ factor\\ (at train time) \end{tabular}} & $\times$1 & 0.911 / 0.907 & 0.898 / 0.887 & 0.849 / 0.836 & 0.783 / 0.776 & 0.706 / 0.701 & 0.645 / 0.639 & 0.593 / 0.582 & 0.553 / 0.541  \\ \cline{3-10}

& $\times$2 & 0.892 / 0.890  & 0.917 / 0.912 & 0.860 / 0.858 & 0.784 / 0.779 & 0.725 / 0.722 & 0.688 / 0.681 & 0.634 / 0.629 & 0.580 / 0.572  \\ \cline{3-10}

 & $\times$3 & 0.869 / \textbf{0.872}  & 0.910 / 0.908 & 0.916 / 0.911 & 0.864 / 0.859 & 0.796 / 0.793 & 0.760 / 0.758 & 0.714 / 0.709 & 0.671 / 0.668  \\ \cline{3-10}

 & $\times$4 & 0.869 / \textbf{0.871}  & 0.894 / \textbf{0.895} & 0.909 / 0.909 & 0.895 / 0.893 & 0.814 / 0.811 & 0.752 / 0.747 & 0.726 / 0.720 & 0.694 / 0.689  \\ \cline{3-10}

 & $\times$5 & 0.832 / \textbf{0.839}  & 0.849 / \textbf{0.856} & 0.898 / \textbf{0.902} & 0.904 / \textbf{0.906} & 0.889 / \textbf{0.905} & 0.843 / \textbf{0.851} & 0.762 / \textbf{0.773} & 0.714 / \textbf{0.727}  \\ \cline{3-10}

 & $\times$6 & 0.806 / \textbf{0.819}  & 0.811 / \textbf{0.823} & 0.871 / \textbf{0.878} & 0.891 / \textbf{0.900} & 0.893 / \textbf{0.899} & 0.887 / \textbf{0.892} & 0.835 / \textbf{0.846} & 0.762 / \textbf{0.773}  \\ \cline{3-10}

 & $\times$7 & 0.777 / \textbf{0.792}  & 0.792 / \textbf{0.804} & 0.841 / \textbf{0.853} & 0.866 / \textbf{0.872} & 0.870 / \textbf{0.879} & 0.878 / \textbf{0.888} & 0.866 / \textbf{0.880} & 0.807 / \textbf{0.821}  \\ \cline{3-10}

 & $\times$8 & 0.744 / \textbf{0.762}  & 0.788 / \textbf{0.801} & 0.809 / \textbf{0.817} & 0.841 / \textbf{0.852} & 0.858 / \textbf{0.872} & 0.867 / \textbf{0.881} & 0.865 / \textbf{0.883} & 0.856 / \textbf{0.874}  \\ \hline
\end{tabular}
}
    \label{tab:ch4_uncertainty_f1}
\end{table*}

\subsection{Methods for Further Improvements}

We then conducted experiments to examine the effectiveness of the proposed techniques in Sec.~\ref{sec:ch4_improve_methods}. 
Note that the results given in Sec.~\ref{sec:ex_superhuman} 
are obtained by the baseline method that do not employ these methods. Thus, the improvements shown in what follows are gained over the baseline method. 

\begin{figure}[thb]
    \centering
    \begin{subfigure}{.32\linewidth}
        \centering
        \includegraphics[width=\linewidth]{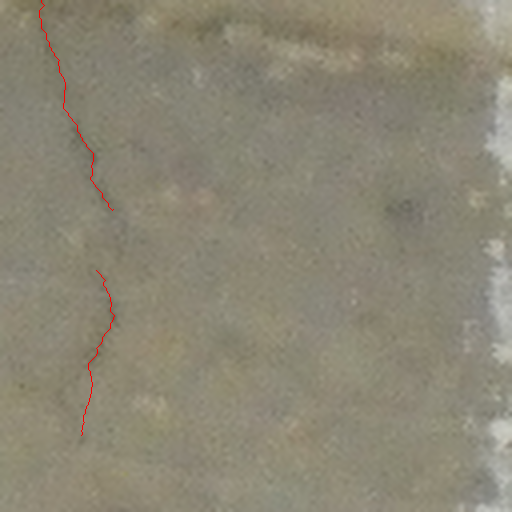}
    \end{subfigure}\hfil%
    \begin{subfigure}{.32\linewidth}
        \centering
        \includegraphics[width=\linewidth]{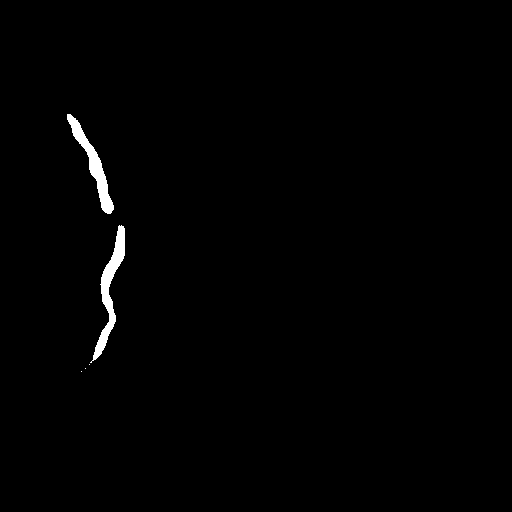}
    \end{subfigure}\hfil%
    \begin{subfigure}{.32\linewidth}
        \centering
        \includegraphics[width=\linewidth]{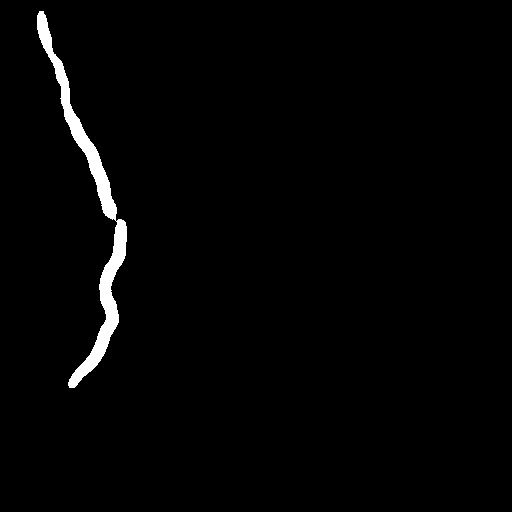}
    \end{subfigure}\\



    \begin{subfigure}{.32\linewidth}
        \centering
        \includegraphics[width=\linewidth]{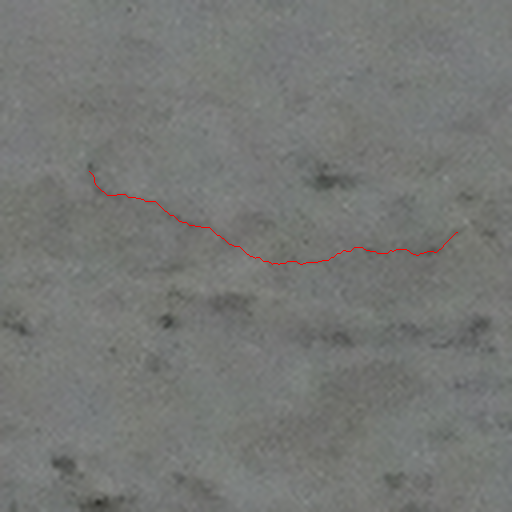}
        \caption{}
    \end{subfigure}\hfil%
    \begin{subfigure}{.32\linewidth}
        \centering
        \includegraphics[width=\linewidth]{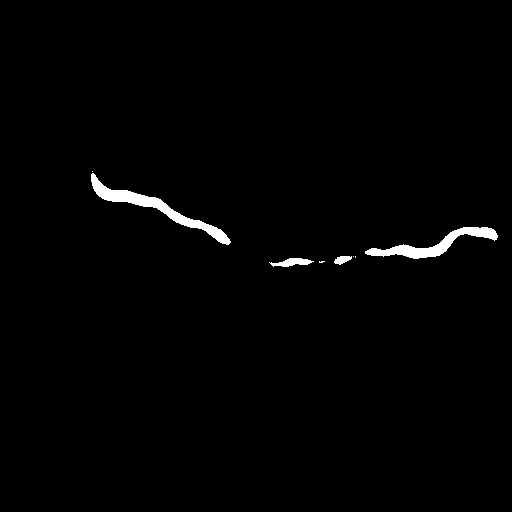}
        \caption{}
    \end{subfigure}\hfil%
    \begin{subfigure}{.32\linewidth}
        \centering
        \includegraphics[width=\linewidth]{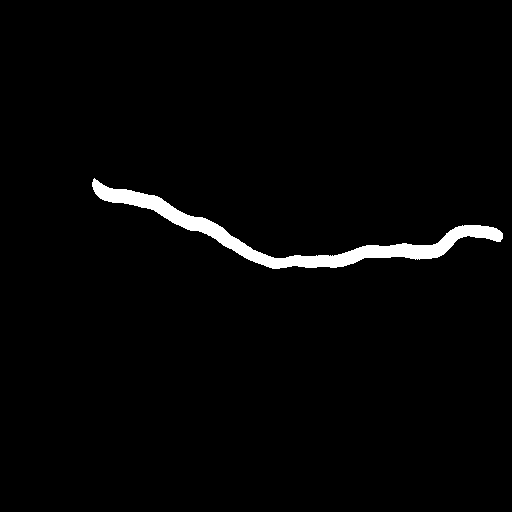}
        \caption{}
    \end{subfigure}

\caption{Improvements obtained by 
$\mathcal{P}$-pooling. We train our CNN on images down- and up-sampled by a factor of 4 (shown in (a)), and test it on original resolution images. Results obtained (b) without and (c) with $\mathcal{P}$-pooling. }
\label{fig:ch5_gem_viz}
\end{figure}

\paragraph{$\mathcal{P}$-pooling}
We first test the effectiveness of $\mathcal{P}$-pooling. We compare the original U-Net*-BN with that with all the pooling layer replaced with $\mathcal{P}$-pooling. We empirically found that $p=3$ works the best. Table~\ref{tab:ch4_p-pooling_f1} show the results obtained by two CNNs with and without $\mathcal{P}$-pooling. It is seen that the pooling method improves accuracy for almost all of pairs of training/test resolutions. The improvements are more significant for the cases of testing on higher-resolution images than training images. Figure \ref{fig:ch5_gem_viz} shows a few examples.

%

\begin{figure}[thb]
    \centering

    \begin{subfigure}{.32\linewidth}
        \centering
        \includegraphics[width=\linewidth]{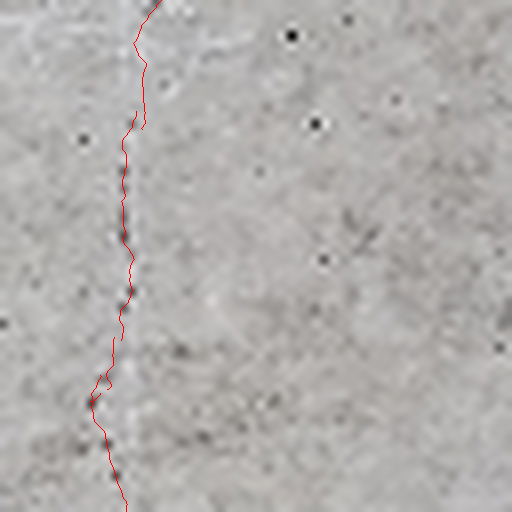}
    \end{subfigure}\hfil%
    \begin{subfigure}{.32\linewidth}
        \centering
        \includegraphics[width=\linewidth]{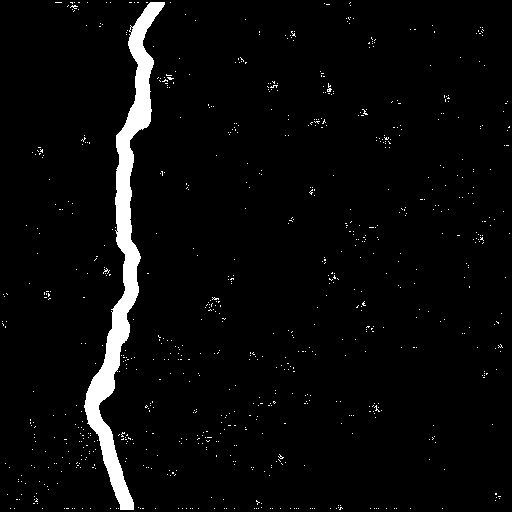}
    \end{subfigure}\hfil%
    \begin{subfigure}{.32\linewidth}
        \centering
        \includegraphics[width=\linewidth]{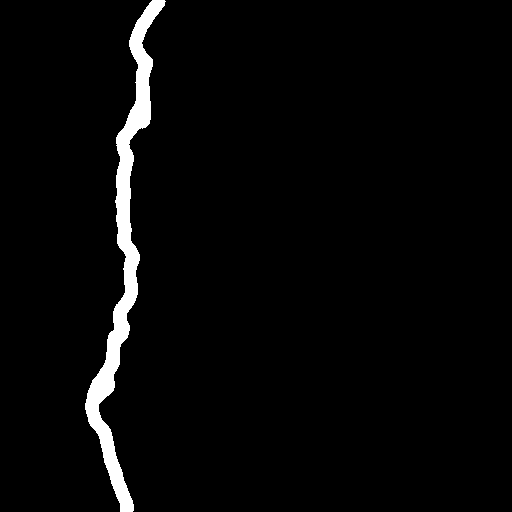}
    \end{subfigure}\\


    \begin{subfigure}{.32\linewidth}
        \centering
        \includegraphics[width=\linewidth]{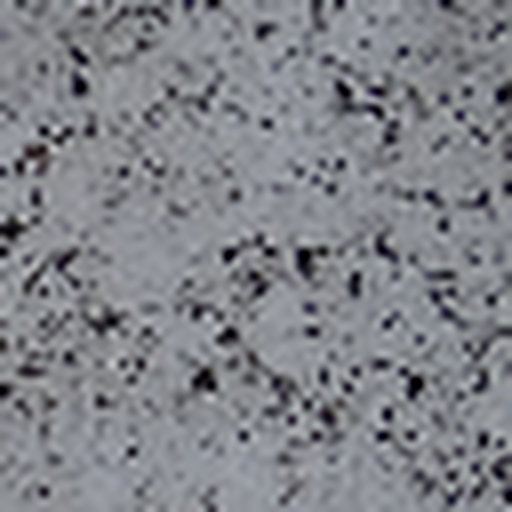}
        \caption{}
    \end{subfigure}\hfil%
    \begin{subfigure}{.32\linewidth}
        \centering
        \includegraphics[width=\linewidth]{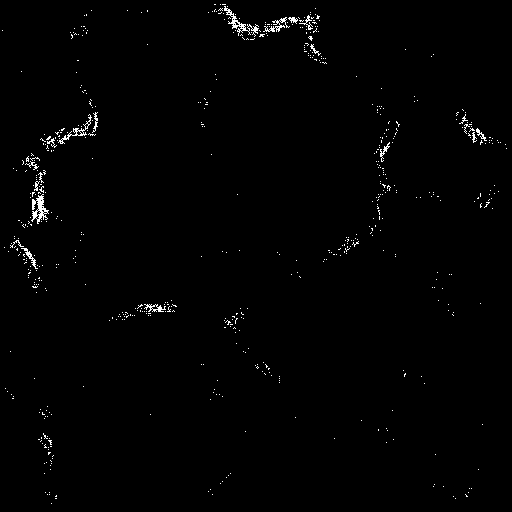}
        \caption{}
    \end{subfigure}\hfil%
    \begin{subfigure}{.32\linewidth}
        \centering
        \includegraphics[width=\linewidth]{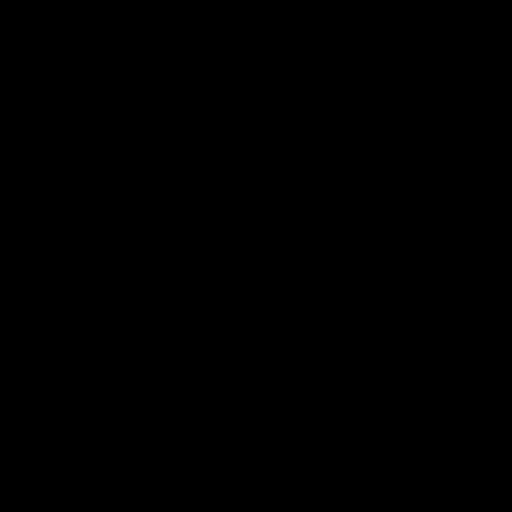}
        \caption{}
    \end{subfigure}


\caption{Improvements by removal of short-segment cracks using learned shape prior. We train our CNN on images down- and up-sampled by a factor of 8 (shown in (a)) and test it on the same resolution images. Results obtained (b) without and (c) with the removal. }
\label{fig:ch4_scn_viz}
\end{figure}

\paragraph{Elimination of Short Segments by Shape Prior}
We then test the effectiveness of the proposed method for elimination of short segments in the post-processing step. We apply the method to the output map of U-Net*-BN with $\mathcal{P}$-pooling. Table~\ref{tab:ch4_sr_f1} shows the accuracy before and after the application of the method. Interestingly, it is seen from the table that it brings about larger improvements for the case of training and testing on the same resolution. Figure \ref{fig:ch4_scn_viz} shows a few examples. The improvements are small when training and testing on different resolutions.

%

\paragraph{Training with uncertainty modeling} 

We train the same network as above using the augmented loss function (\ref{eqn:uncertainty_loss}), followed by the elimination of short segments. Table \ref{tab:ch4_uncertainty_f1} shows comparisons between the results obtained with the MSE loss and those obtained with the new loss. The results show that the employment of the uncertainty model improves only the cases of training the CNN on low resolution images. It even worsen the detection accuracy for the CNNs trained on high resolution images. We should use this method selectively according to the chosen pair of training/test resolutions.

\section{Summary and Conclusion}

We have presented several approaches to improve the accuracy of the detection of cracks. Stating that the difficulty with crack detection lies in detecting thin cracks, we first presented a formulation of the problem that is more proper than the standard one in the literature. We then proposed a method for obtaining super-human labels by controlling the resolution of training images. We also presented three methods aiming at further improvement of the accuracy of thin crack detection. The experimental results show the effectiveness of the proposed approaches. 



{\small
\bibliographystyle{ieee_fullname}
\bibliography{egbib}
}

\end{document}